\documentclass[10pt,twocolumn,letterpaper]{article}

\usepackage{iccv}
\usepackage{times}
\usepackage{epsfig}
\usepackage{graphicx}
\usepackage{amsmath}
\usepackage{amssymb}
\usepackage{multirow}
\usepackage{subfig} 

\usepackage[pagebackref=true,breaklinks=true,letterpaper=true,colorlinks,bookmarks=false]{hyperref}

\iccvfinalcopy 

\newcommand{\better}[1]{\textcolor[rgb]{0, 0.8, 0.2}{#1}}
\newcommand{\worse}[1]{\textcolor[rgb]{0.8, 0.2, 0}{#1}}
\newcommand{\second}[1]{\underline{#1}}

\begin{document}

\title{MBPTrack: Improving 3D Point Cloud Tracking with Memory Networks and Box Priors}

\author{Tian-Xing Xu$^{1}$\quad Yuan-Chen Guo$^{1}$\quad Yu-Kun Lai$^{2}$ \quad Song-Hai Zhang$^{1}$\\
$^{1}$ Tsinghua University, China \quad $^{2}$ Cardiff University, United Kingdom\\
{\tt\small $^{1}$\{xutx21@mails., guoyc19@mails.,  shz@\}tsinghua.edu.cn\quad $^{2}$LaiY4@cardiff.ac.uk}
}

\maketitle

\begin{abstract}
3D single object tracking has been a crucial problem for decades with numerous applications such as autonomous driving. Despite its wide-ranging use, this task remains challenging due to the significant appearance variation caused by occlusion and size differences among tracked targets. To address these issues, we present \textbf{MBPTrack}, which adopts a \textbf{M}emory mechanism to utilize past information and formulates localization in a coarse-to-fine scheme using \textbf{B}ox \textbf{P}riors given in the first frame. Specifically, past frames with targetness masks serve as an external memory, and a transformer-based module propagates tracked target cues from the memory to the current frame. To precisely localize objects of all sizes, MBPTrack first predicts the target center via Hough voting. By leveraging box priors given in the first frame, we adaptively sample reference points around the target center that roughly cover the target of different sizes. Then, we obtain dense feature maps by aggregating point features into the reference points, where localization can be performed more effectively. Extensive experiments demonstrate that MBPTrack achieves state-of-the-art performance on KITTI, nuScenes and Waymo Open Dataset, while running at 50 FPS on a single RTX3090 GPU.  
\end{abstract}

\section{Introduction}

The ability to track objects in 3D space is essential for numerous applications, including robotics~\cite{budiharto2020design,jiang2021high}, autonomous driving~\cite{yin2021center,kuang2020probabilistic}, and surveillance systems~\cite{thys2019fooling}. Given the initial state of a specific object, the aim of 3D single object tracking (SOT) is to estimate the pose and position of the tracked target in each frame. Early approaches~\cite{spinello2010layered,luber2011people,pieropan2015robust} rely heavily on RGB information, which often struggle to handle changing lighting conditions. Therefore, recent research works~\cite{giancola2019leveraging,qi2020p2b,hui20213d,hui20223d,zheng2022beyond,xu2022cxtrack} have focused on using point clouds to solve 3D object tracking for their unique advantages, such as accurate spatial information and robustness to illumination changes. 

\begin{figure}
\centering
\includegraphics[width=1.0\linewidth]{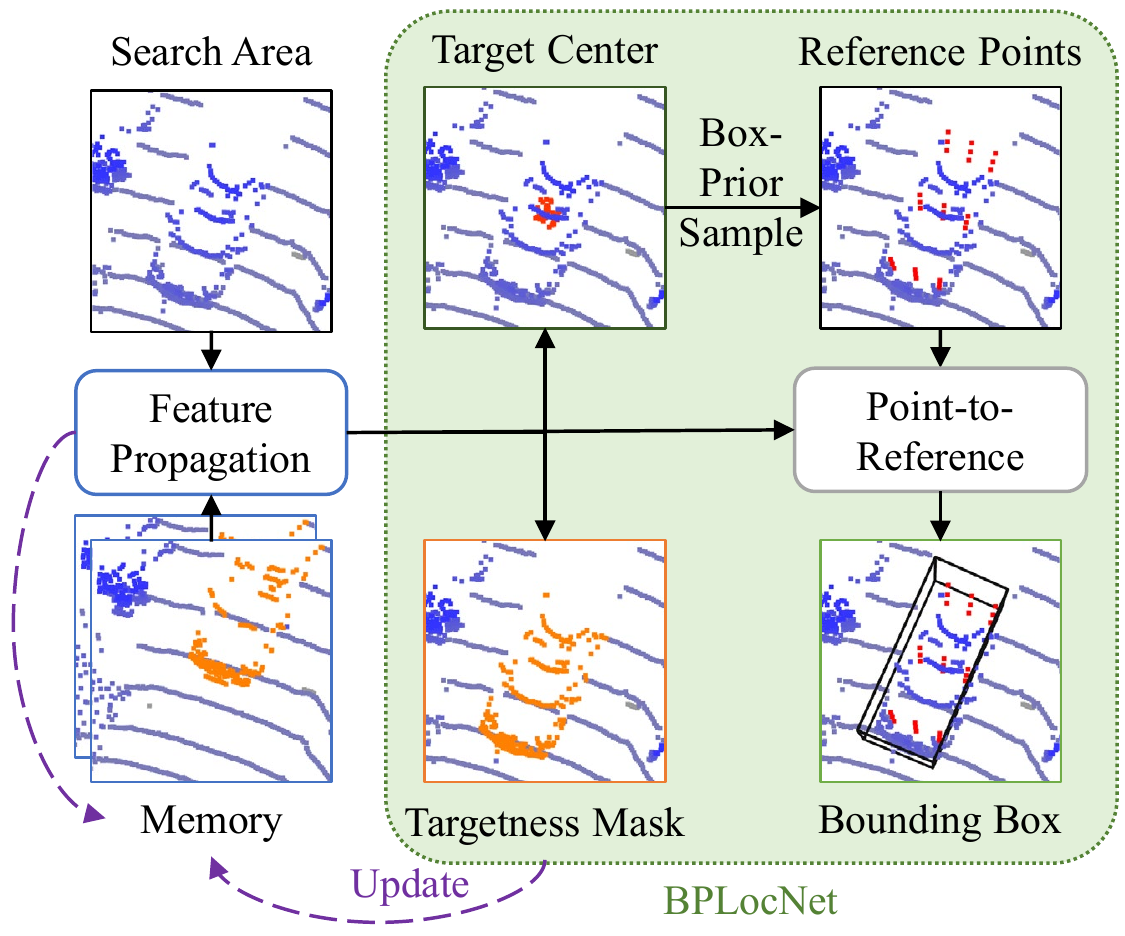}
\vspace{-0.5cm}
\caption{\textbf{An illustration of our proposed MBPTrack for 3D SOT task.} MBPTrack employs a memory mechanism to propagate target cues from historical frames and then utilizes a localization head, named BPLocNet, for coarse-to-fine bounding box prediction. BPLocNet first samples reference points around each predicted target center using a bounding box prior, which adaptively covers the tracked targets of all sizes. Then, BPLocNet aggregates local features from reference points for further refinement.  }
\label{fig:paradigm}
\vspace{-0.5cm}
\end{figure}

Existing methods~\cite{qi2020p2b,wang2021mlvsnet,zheng2021box,shan2021ptt,cui20213d,hui20213d,zhou2022pttr,hui20223d} for the 3D SOT task predominantly follow the Siamese paradigm, which takes the target template cropped from the previous frame and search area in the current frame as input, and then localizes the target in an end-to-end manner using a localization network such as Region Proposal Network~\cite{qi2019deep} (RPN). Different from previous methods, M2-Track~\cite{zheng2022beyond} explicitly models the target's motion between two successive frames and CXTrack~\cite{xu2022cxtrack} proposes to exploit the spatial contextual information across adjacent frames. Despite achieving promising results on popular datasets, the aforementioned methods propagate target cues solely from the latest frame to the current frame, thereby neglecting rich information contained in other past frames. This limitation renders 3D SOT a challenging task, especially in cases of large appearance variation or target disappearance caused by occlusion. To this end, TAT~\cite{lan2022tat} exploits temporal information by sampling a set of high-quality target templates cropped from historic frames for reliable target-specific feature propagation. However, neglecting information in the latest frame could result in the network failing to capture lasting appearance changes, such as the gradual sparsification of point clouds as the tracked target moves further away. TAT also
ignores the contextual information around the target, which is essential for 3D SOT~\cite{xu2022cxtrack}, thereby leading to limited tracking performance.  

In addition, the substantial differences in size and geometry across the various categories of tracked targets also pose challenges for 3D SOT, which has been overlooked by previous works. The localization networks adopted in existing methods can be categorized into two paradigms, namely point-based~\cite{zhou2022pttr,xu2022cxtrack,qi2020p2b} and voxel-based~\cite{hui20213d}. For voxel-based localization heads like V2B~\cite{hui20213d}, tracked targets with simple shapes and large sizes such as vehicles, can fit well in voxels, leading to more precise localization than point-based heads such as X-RPN~\cite{xu2022cxtrack}. However, for categories such as pedestrians, which have complex geometries and small sizes, voxelization leads to considerable information loss, thereby degrading tracking performance. As mentioned in V2B~\cite{hui20213d}, the choice of different voxel sizes can significantly impact tracking performance.

To address the above issues, we present MBPTrack, a memory-based network for the 3D SOT task. Our approach relies on a memory mechanism to leverage rich spatial and temporal contextual information in historical frames and utilizes bounding box priors to address the challenge of size differences among tracked targets. Specifically, past frames with targetness masks serve as an external memory, and we draw inspiration from DeAOT~\cite{yang2022decoupling}, which has achieved great success in video object segmentation, to design a transformer-based module that propagates information from this memory to the current frame. It further decouples geometric features and targetness features into two processing branches with shared attention maps to enable effective learning of geometric information. Unlike TAT~\cite{lan2022tat}, MBPTrack fully utilizes both spatial and temporal contextual information around the target without cropping or sampling, thereby handling appearance variation and target disappearance/reappearance better than previous works. 
To achieve accurate localization of targets of different sizes, we introduce BPLocNet, a coarse-to-fine localization network that captures size information by leveraging the bounding box given in the first frame. BPLocNet first predicts the potential target centers as well as the targetness mask used to update the memory mechanism. We adopt a box-prior sampling method to sample reference points around the predicted target centers, adaptively covering the tracked target. Then, we aggregate point-wise features into the reference points, to obtain a dense feature map with spatial information, which is fed into a 3D CNN to predict precise bounding boxes. Extensive experiments demonstrate that MBPTrack outperforms existing methods by a large margin on three benchmark datasets, while running at 50 FPS on a single NVIDIA RTX3090 GPU. Furthermore, we demonstrate that using our proposed localization network in existing frameworks can consistently improve tracking accuracy.

In summary, our main contributions are as follows:

\begin{itemize}
    \item To the best of our knowledge, we are the first to exploit both spatial and temporal contextual information in the 3D SOT task using a memory mechanism.    
    \item We propose a localization network that utilizes box priors to localize targets of different sizes in a coarse-to-fine manner, which is shown to be effective in various 3D SOT frameworks.
    \item Experimental results demonstrate that MBPTrack outperforms existing methods, achieving state-of-the-art online tracking performance.

\end{itemize}

\section{Related Work}

As the pioneering work for point cloud-based 3D SOT, SC3D~\cite{giancola2019leveraging} computes feature similarity between the target template and a potentially large number of candidate proposals, which are sampled by Kalman filter in the search area. However, the heuristic sampling is time-consuming, and the pipeline cannot be end-to-end trained. To balance performance and efficiency, P2B~\cite{qi2020p2b} adopts a Region Proposal Network~\cite{qi2019deep} to generate high-quality 3D proposals. 
The proposal with the highest score is selected as the final output. Many follow-up works adopt the same paradigm. MLVSNet~\cite{wang2021mlvsnet} enhances P2B by performing multi-level Hough voting for effectively aggregating information at different levels. BAT~\cite{zheng2021box} designs a box-aware feature fusion module to capture the explicit part-aware structure information. V2B~\cite{hui20213d} proposes to transform point features into a dense bird's eye view feature map to tackle the sparsity of point clouds. LTTR~\cite{cui20213d}, PTTR~\cite{zhou2022pttr}, CMT~\cite{guo2022cmt} and STNet~\cite{hui20223d} introduce various attention mechanisms into the 3D SOT task for better target-specific feature propagation. PTTR~\cite{zhou2022pttr} also proposes a light-weight Prediction Refinement Module for coarse-to-fine localization. However, these methods rely wholly on the appearance of the target, so tend to drift towards distractors in dense traffic scenes~\cite{zheng2022beyond}. To this end, M2-Track~\cite{zheng2022beyond} introduces a motion-centric paradigm that explicitly models the target's motion between two adjacent frames. CXTrack~\cite{xu2022cxtrack} exploits contextual information across adjacent frames to improve tracking results. Although achieving promising results, these methods only exploit the target cues in the latest frame. The overlook of rich information in historical frames may hinder precise localization in the case of large appearance variation or target disappearance caused by occlusion. 

TAT~\cite{lan2022tat} is the first work to exploit the rich temporal information. It samples high-quality target templates from historical frames and adopts an RNN-based module~\cite{chung2014gru} to aggregation target cues from multiple templates. However, the overlook of low-quality target templates in the latest frame makes the network fail to capture lasting appearance variation caused by long-term partial occlusion. It also ignores the spatial contextual information in the historical frames, which is essential for 3D SOT, as mentioned in CXTrack~\cite{xu2022cxtrack}. Besides, none of the aforementioned methods consider the size differences of tracked objects. For example, compared with pedestrian, vehicles have simple shapes and large sizes, which fit well in voxels. Thus voxel-based networks such as STNet~\cite{hui20223d} achieve better performance on the Car category than point-based networks like CXTrack~\cite{xu2022cxtrack}, but face great challenges on the Pedestrian category.  We argue that object occlusion and size difference are two main factors that pose great challenges for 3D SOT.

\section{Method}
\subsection{Problem Definition}

Given the 3D bounding box (BBox) of a specific target in the first frame, 3D SOT aims to localize the target by predicting its bounding box in subsequent frames. The frame at timestamp $t$ is represented as a point cloud $\mathcal{P}_t\in \mathbb{R}^{\dot{N}_{t} \times 3}$, where $\dot{N}_t$ is the number of points. The 3D BBox $\mathcal{B}_t\in \mathbb{R}^7$ at timestamp $t$ is parameterized by its center ($xyz$ coordinates), orientation (heading angle $\theta$ around the up-axis) and size (width $w$, length $l$ and height $h$). Even for non-rigid objects like pedestrians, the size of the tracked target remains approximately unchanged in 3D SOT. Thus, for each frame $\mathcal{P}_t$, we only regress the translation offset ($\Delta x_t, \Delta y_t, \Delta z_t$) and the rotation angle ($\Delta \theta_t$) from $\mathcal{P}_{t-1}$ to $\mathcal{P}_t$ to simplify the tracking task, with access to historical frames $\{\mathcal{P}_i\}_{i=1}^{t}$. The 3D BBox $\mathcal{B}_t$ can be easily obtained by applying a rigid body transformation to $\mathcal{B}_{t-1}$ from the previous frame. Additionally, to indicate a more precise location of the tracked target at timestamp $t$, we predict a targetness mask $\mathcal{\dot{M}}_{t} = (m^1_{t}, m^2_{t}, \cdots, m^{\dot{N}_{t}}_{t}) \in \mathbb{R}^{\dot{N}_{t}}$ frame by frame, where the mask $m_{t}^i$ represents the possibility of the $i$-th point $p_{t}^i\in \mathcal{P}_t$ being within $\mathcal{B}_t$ ($\mathcal{\dot{M}}_{1}$ is computed using the given $\mathcal{B}_1$). Hence, we can formulate 3D SOT at timestamp $t(t>1)$ as learning the following mapping
\begin{multline}
    \mathcal{F}(\{\mathcal{P}_{i}\}_{i=1}^{t-1},\{\dot{\mathcal{M}}_i\}_{i=1}^{t-1}, \mathcal{P}_{t}, \mathcal{B}_1) \mapsto \\
(\Delta x_t, \Delta y_t, \Delta z_t, \Delta \theta_t, \dot{\mathcal{M}}_t)
\label{eq:paradigm}
\end{multline}

\subsection{Overview}

\begin{figure*}
\centering
\includegraphics[width=1.0\linewidth]{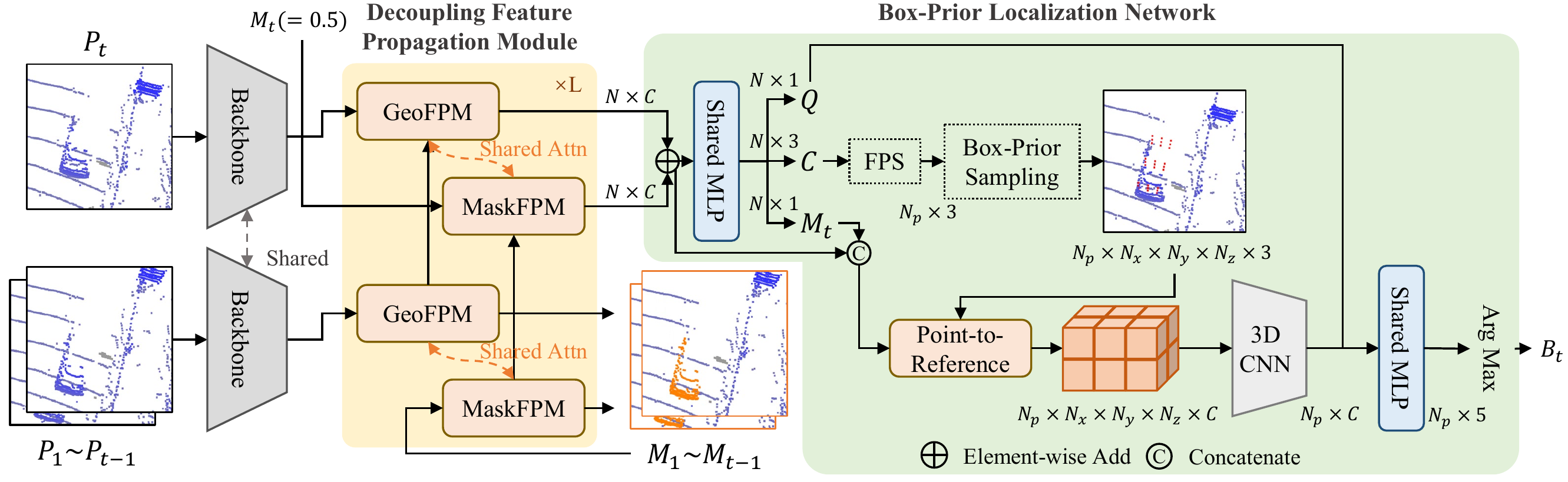}
\vspace{-0.6cm}
\caption{\textbf{An overview of our proposed MBPTrack architecture.} We employ a backbone to extract geometric features. Then, the past frames with their targetness mask serve as an external memory and the decoupling feature propagation module is used to propagate rich target cues from historical frames. To tackle the size difference problem, we propose a coarse-to-fine network, named BPLocNet, for object localization, which leverages box priors to sample reference points that adaptively cover the target of different sizes for precise localization. }
\label{fig:overview}
\vspace{-0.5cm}
\end{figure*}

Following Eq.~\ref{eq:paradigm}, we design a memory-based framework, MBPTrack, to capture the spatial and temporal information in the historical frames and tackle the size difference across various categories of tracked targets. As illustrated in Fig.~\ref{fig:overview}, given an input sequence $\{\mathcal{P}_{i}\}_{i=1}^t$ of a dynamic 3D scene, we first employ a shared backbone to embed the local geometric information in each frame into point features, denoted by $\mathcal{X}_i \in \mathbb{R}^{{N}\times C}$ for the $i$-th frame. Here $N$ is the number of point features and $C$ denotes the number of feature channels. The corresponding targetness masks $\mathcal{M}_i\in\mathbb{R}^{N\times 1}(i<t)$ are obtained from $\dot{\mathcal{M}}_i$ (either from the first frame or estimated from past frames) to identify the tracked target in past frames. The targetness mask $\mathcal{M}_t$ for the current frame is initialized with 0.5 as it is unknown. Then, we design a transformer-based decoupling feature propagation module (DeFPM, Sec.~\ref{sec:dfpm}) to leverage both temporal and spatial context present in the dynamic 3D scene. Finally, we develop a simple yet efficient localization network, BPLocNet, which formulates the localization of targets as coarse-to-fine prediction using box priors to tackle size differences among tracked targets.   

\subsection{Decoupling Feature Propagation Module}
\label{sec:dfpm}

\begin{figure}
\centering
\includegraphics[width=1.0\linewidth]{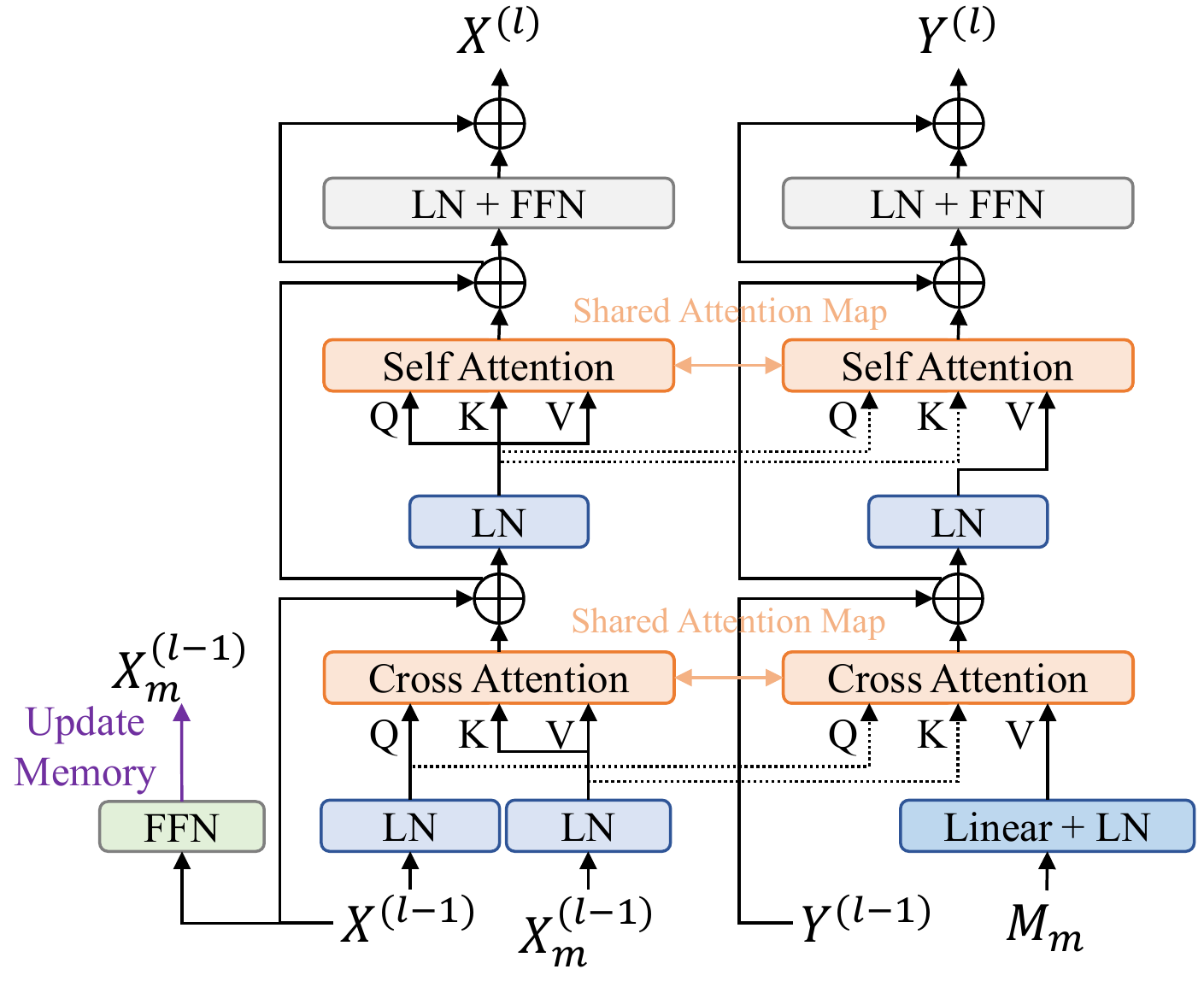}
\vspace{-0.7cm}
\caption{\textbf{Decoupling Feature Propagation Module (DeFPM).} DeFPM decouples the propagation of geometric information and targetness information into two branches to avoid the loss of geometric information into deep propagation layers. Both branches have the same hierarchical structure with shared attention maps. }
\label{fig:dfpm}
\vspace{-0.5cm}
\end{figure}

Inspired by the success of DeAOT~\cite{yang2022decoupling} in video object segmentation, we introduce a decoupling feature propagation module (DeFPM) into the 3D SOT task, which relies on a memory mechanism to explore both spatial and temporal information from the past frames while propagating target cues into the current frame. Previous work~\cite{yang2022decoupling} indicates that integrating targetness information will inevitably cause the loss of object-agnostic geometric information. Hence, DeFPM decouples the propagation of geometric features and mask features to learn more distinct geometric embeddings, which is essential for handling sparse point clouds. DeFPM consists of $N_L=2$ identical layers with two parallel branches, as illustrated in Fig.~\ref{fig:dfpm}. Each layer includes three main parts, \textit{i}.\textit{e}., a cross-attention module that propagates both target cues and temporal context from past frames to the current frame, a self-attention module that captures long-range contextual information in the current frame, and a feed-forward network for feature refinement. 

To formulate the feature propagation from the memory to the current frame, we first define the input of the $l$-th layer. Let $X^{(l-1)}\in \mathbb{R}^{N\times C}$ and $X_\text{m}^{(l-1)} \in \mathbb{R}^{TN\times C}$ denote the geometric features in the current frame and from the memory, where $T$ represents the memory size (the number of memory frames). We adopt a ``pre-norm'' transformer design~\cite{misra2021end}, which employs a layer normalization~\cite{ba2016layer} operation $\text{LN}(\cdot)$ before the attention mechanism, written as 
\begin{align}
    \overline{X} &= \text{LN}(X^{(l-1)}) \\
    \overline{X}_{M} &= \text{LN}(X^{(l-1)}_M)    
\end{align}
The attention mechanism~\cite{vaswani2017attention} is the basic block of our proposed DeFPM, which takes the query $Q\in \mathbb{R}^{n\times d}$, key $K\in \mathbb{R}^{n\times d}$ and value $V\in \mathbb{R}^{n\times d}$ as input, and then computes the similarity matrix between the query and the key to obtain a weighted sum of $V$
\begin{align}
    \text{Attn}(Q,K,V) = \text{softmax}(\frac{QK^T}{\sqrt{d}})V.
\end{align}
Notably, we add positional embeddings of the coordinates to the query and key, which is omitted from the formula of attention for brevity. 
Hence, the cross attention operation of the geometric branch can be formulated as 
\begin{align}
  \widetilde{X} &= X^{(l-1)}+\text{Attn}(\overline{X}W^c_Q, \overline{X}_{M}W^c_K, \overline{X}_{M}W^c_{V,G}),
\end{align}
where $W^c_Q\in \mathbb{R}^{C\times d}$, $W^c_K\in \mathbb{R}^{C\times d}$ and $W^c_{V,G}\in \mathbb{R}^{C\times d}$ are learnable parameter matrices and $d$ is the channel dimension. 
To update the memory bank, we adopt a lightweight feed-forward network on the input $X^{(l-1)}$ to obtain the reference features of the current frame, which ensures the effectiveness and efficiency of the memory mechanism. 

The mask branch is designed to propagate targetness information into the current frame to indicate the tracked target, which shares the attention maps with the geometric branch. Suppose $Y^{(l-1)}\in \mathbb{R}^{N\times C}$ denotes the mask features output by the $(l-1)$-th layer ($Y^{(0)} = \phi(\mathcal{M}_t)$, where $\phi$ is a linear projection layer) and $M_m\in \mathbb{R}^{TN\times 1}$ denotes the targetness masks of all the frames saved in memory. We project the input masks $M_m$ to mask embeddings using a shared linear transformation $\varphi(\cdot)$ with a layer normalization
\begin{align}
    \overline{Y_m} = \text{LN}(\varphi(M_m))
\end{align}
The output of the cross-attention operation is given by
\begin{align}
    \widetilde{Y} &= Y^{(l-1)}+\text{Attn}(\overline{X}W^c_Q, \overline{X}_{M}W^c_K, \overline{Y}_{M}W^c_{V,M}),
\end{align}

To explore the contextual information within the current frame and enhance the point features, DeFPM subsequently employs a global self-attention operation, which can be formulated similarly to the cross-attention operation:
\begin{align}
    \widehat{X} &= \widetilde{X}+\text{Attn}(\dot{X}W^s_Q, \dot{X}W^s_K, \dot{X}W^s_{V,G})\\
    \widehat{Y} &= \widetilde{Y}+\text{Attn}(\dot{X}W^s_Q, \dot{X}W^s_K, \dot{Y}W^s_{V,G})\\
    \text{where}\;\;  \dot{X} &= \text{LN}(\widetilde{X}), \dot{Y} = \text{LN}(\widetilde{Y})
\end{align}

Finally, two fully connected feed-forward networks are used to separately refine the point features and mask features, which can be written as 
\begin{align}
    X^{(l)} &= \widehat{X} + \text{FFN}(\text{LN}(\widehat{X})) \\
    Y^{(l)} &= \widehat{Y} + \text{FFN}(\text{LN}(\widehat{Y})) \\
    \text{where}\;\; \text{FFN}(x) &= \text{ReLU} ( xW_1+b_1)W_2+b_2
\end{align}

\subsection{Box-Prior Localization Network}

The difference in size and geometry among the tracked targets poses great challenges to existing localization networks. To address the above concern, we design a novel localization network, named BPLocNet, that formulates the localization of the target in a coarse-to-fine manner using box priors given in the first frame, as illustrated in Fig.~\ref{fig:overview}. Previous works~\cite{qi2020p2b,zhou2022pttr,hui20213d,hui20223d} mainly use the bounding box to crop the target template from previous frames while ignoring the size information about the target. Hence, we propose to adaptively sample reference points that roughly cover the targets of different sizes using the given bounding box, and then refine the prediction for precise localization.

\begin{figure}
\centering
\includegraphics[width=1.0\linewidth]{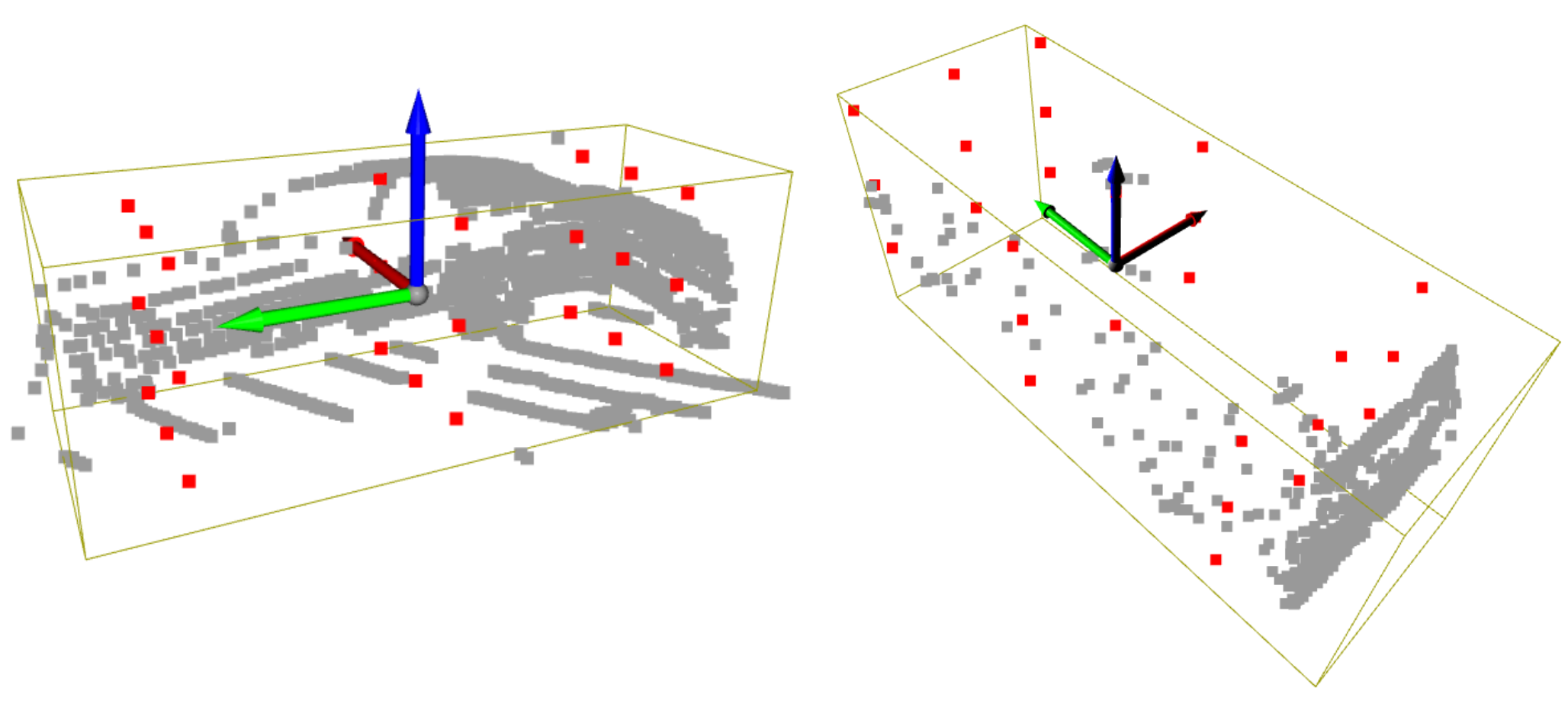}
\vspace{-0.7cm}
\caption{\textbf{Box-Prior Sampling.} Red indicates the reference points that roughly cover the tracked targets. We sample 3 points along each axis for visualization. }
\label{fig:sample}
\vspace{-0.5cm}
\end{figure}

\noindent\textbf{Box-prior sampling.} We apply a shared MLP on the fused point features $F=X^{(N_L)} + Y^{(N_L)}\in \mathbb{R}^{N\times C}$ to predict the potential target center $\mathcal{C}\in \mathbb{R}^{N\times 3}$ via Hough voting, as well as a point-wise targetness mask $\mathcal{M}_t$ for memory update. Each target center prediction can be viewed as a proposal center, while we use further point sampling to sample a subset $\mathcal{C}_p$ in $\mathcal{C}$ to be of size $N_p$ for efficiency. Suppose $w,l,h$ denote the width, length and height of the 3D bounding box $\mathcal{B}_1$ given in the first frame along each axis. Leveraging the proposal centers $\mathcal{C}_p$ and the size information $w,l,h$, we can sample a set of reference points $\mathcal{R}_c$ for each center $c\in \mathcal{C}_p$ (as shown in Fig. ~\ref{fig:sample}), which can be formulated as follows
\small{
\begin{multline}
    \mathcal{R}_c = \bigg\{c+s_{i,j,k}\bigg |s_{i,j,k} = \bigg(\frac{2i-n_x-1}{2n_x}w, \frac{2j-n_y-1}{2n_y}l,\\ 
 \frac{2k-n_z-1}{2n_z}h \bigg)  \;\;\forall i\in [1, n_x], j\in [1, n_y], k\in [1, n_z]\bigg\}
\end{multline}}
\normalsize
where $n_x, n_y, n_z$ denote the numbers of samples along axes. All reference points form a point set $\mathcal{R} =  \bigcup_{c \in \mathcal{C}_p} \mathcal{R}_c$. Despite the simplicity, reference points have the following properties, which can benefit the localization task:
\begin{itemize}
    \item \textbf{Coarse prediction.} We observe the rotation angles $\Delta \theta$ of tracked targets between two consecutive frames are small in most cases, especially for vehicle tracking. Hence, reference points can serve as a coarse prediction of the localization of targets.
    \item \textbf{Adaptive to different sizes.} By leveraging size information, reference points are uniformly distributed in the bounding box of tracked targets, making the network adaptive to targets of different sizes.
    \item \textbf{Shape prior.} For targets such as vehicles that have simple shapes, reference points provide a strong shape prior, which roughly cover the targets in 3D space.
\end{itemize}

\noindent\textbf{Point-to-reference feature transformation.} Due to the fixed relative position of reference points, we can obtain a 3D dense feature map from unordered point features, where the localization can be performed more effectively. We first integrate the targetness mask score $m_j\in \mathcal{M}_t$ into the point features $f_j\in F$ using a shared MLP $h(\cdot)$ \begin{align}
    \widehat{f}_j = h([f_j;m_j]) 
\end{align}
where $[\cdot;\cdot]$ is the concatenation operation. Then, we adopt a modified EdgeConv~\cite{wang2019dynamic} operator to aggregate information from the neighborhood points $j\in \mathcal{N}(r)$ with features $\widehat{f}_j \in F$ to the reference point $r\in \mathcal{R}$, written as follows
\begin{align}
    f_r = \max_{j:j\in \mathcal{N}(r)} e([\widehat{f}_j;x_j-r;r])
\end{align}
where $e(\cdot)$ denotes a shared MLP. We arrange the features $f_r$ in a predefined order with respect to the coordinates of reference points to generate the 3D dense feature map $\mathcal{Z}\in \mathbb R^{n_x\times n_y \times n_z \times C}$ for each proposal $c\in \mathcal{C}$. Finally, the 3D feature maps $\mathcal{Z}$ are fed into a shared 3D CNN to obtain proposal-wise features.  

\noindent\textbf{Coarse-to-fine score prediction.} As mentioned in M2-Track~\cite{zheng2022beyond}, distractors are widespread in dense traffic scenes. 3D CNN mainly captures the appearance of the target, while failing to distinguish the target from distractors. Thus, we also use the point features output by DeFPM to predict a quality score $\mathcal{Q}\in \mathbb{R}^{N\times 1}$, responsible for measuring the distance between the predicted target center $\mathcal{C}$ and the ground truth. We integrate the quality score into the proposal-wise features to obtain the box parameters $\mathcal{B}^{N_p\times 4}$ with refined targetness scores $S\in \mathbb{R}^{N_p\times 1}$.

\subsection{Implementation Details}

\noindent \textbf{Loss functions.} Our proposed MBPTrack is trained in an end-to-end manner. The predicted targetness mask $\mathcal{M}_t$ is supervised by a standard cross-entropy loss, denoted as $\mathcal{L}_\text{m}$. For the target center prediction, we use an MSE (Mean Squared Error) loss $\mathcal{L}_\text{c}$. Following P2B~\cite{qi2020p2b}, we consider predicted centers near the ground truth target center (\textless 0.3m) as positive and others as negative to obtain the ground truth of quality score $Q$ and targetness score $S$, which are supervised by cross-entropy loss $\mathcal{L}_\text{q}$ and $\mathcal{L}_\text{s}$, respectively. Only the bounding box parameters of positive proposals are supervised via a smooth-L1 loss $\mathcal{L}_\text{bbox}$. The final loss can be written as
\begin{align}
    \mathcal{L} = \lambda_\text{m} \mathcal{L}_\text{m} + \lambda_\text{c} \mathcal{L}_c + \lambda_\text{q} \mathcal{L}_\text{q} + \lambda_\text{s} \mathcal{L}_s + \mathcal{L}_\text{bbox}
\end{align}
where $\lambda_\text{m}$(=0.2), $\lambda_\text{c}$(=10.0), $\lambda_\text{q}$(=1.0) and $\lambda_\text{s}$(=1.0) hyperparameters are used to balance the component losses. 

\noindent \textbf{Positive sampling.} We observe that for objects with complex shapes such as pedestrians, it is difficult to regress precise target centers for all point features. Hence, positive proposals (\textless 0.3m) for box parameter prediction are much fewer than negative proposals. To balance positive and negative proposals, we replace part of the predicted proposal centers with positive centers generated by applying a small perturbation to the ground truth center for nonrigid objects such as pedestrians and cyclists during training.

\noindent \textbf{Training \& Testing.} We sample 8 consecutive frames from the point cloud sequence to form a training sample. MBPTrack only exploits temporal information from the previous 2 frames in training and leverages 3 previous frames during testing to balance efficiency and effectiveness (Sec.~\ref{sec:ablation}). 
MBPTrack reuses the previous prediction if the tracked target is lost ($\max(\mathcal{M}_t)<0.2$) in the current frame. More details can be seen in the supplementary material.

\section{Experiments}

\subsection{Settings}

 \noindent\textbf{Datasets.} We adopt three popular large-scale datasets, namely KITTI~\cite{geiger2012kitti}, NuScenes~\cite{caesar2020nuscenes} and Waymo Open Dataset~\cite{sun2020waymo} (WOD), to validate the effectiveness of our model. KITTI contains 21 video sequences for training and 29 video sequences for testing. Due to the inacessibility of the test labels, we follow previous work~\cite{qi2020p2b} and split the training dataset into three subsets, sequences 0-16 for training,
17-18 for validation, and 19-20 for testing. NuScenes is more challenging than KITTI for its larger data volumes, containing 700/150/150 scenes for training/validation/testing. For WOD, we follow LiDAR-SOT~\cite{zhang2021modelfree} to evaluate our method on 1121 tracklets, which are divided into easy, medium and hard subsets based on the sparsity of point clouds.  

\noindent\textbf{Evaluation metrics.} We follow One Pass Evaluation~\cite{kristan2016novel}. 
For the predicted and ground truth bounding boxes, Success is defined as the Area Under Curve (AUC) for the plot showing the ratio of frames where the Intersection Over Union (IOU) is greater than a threshold, ranging from 0 to 1, while Precision denotes the AUC for the plot showing the ratio of frames where the distance between their centers is within a threshold, ranging from 0 to 2 meters. 

\subsection{Results}

\begin{table}
\begin{center}
\caption{\textbf{Comparisons with the state-of-the-art methods on KITTI dataset}. ``Mean'' is the average result weighted by frame numbers. ``\second{Underline}'' and ``\textbf{Bold}'' denote previous and current best performance, respectively. Success/Precision are used for evaluation.}
\vspace{-0.3cm}
\label{kitti}
\resizebox{\linewidth}{!}{
\setlength{\tabcolsep}{3pt}
\begin{tabular}{c|c|c|c|c|c}
\hline 
\multirow{2}{*}{Method} & Car & Pedestrian & Van & Cyclist & Mean\\
&(6424) & (6088) & (1248) & (308) & (14068) \\
\hline
 SC3D & 41.3/57.9 & 18.2/37.8 & 40.4/47.0 & 41.5/70.4 & 31.2/48.5 \\
 P2B & 56.2/72.8 & 28.7/49.6 & 40.8/48.4 & 32.1/44.7 & 42.4/60.0 \\ 
 3DSiamRPN & 58.2/76.2 & 35.2/56.2 & 45.7/52.9 & 36.2/49.0 & 46.7/64.9 \\
 LTTR & 65.0/77.1 & 33.2/56.8 & 35.8/45.6 & 66.2/89.9& 48.7/65.8 \\
 MLVSNet & 56.0/74.0 & 34.1/61.1 & 52.0/61.4 & 34.3/44.5 & 45.7/66.7 \\ 
 BAT & 60.5/77.7 & 42.1/70.1 & 52.4/67.0 & 33.7/45.4 & 51.2/72.8 \\
 PTT & 67.8/81.8 & 44.9/72.0 & 43.6/52.5 & 37.2/47.3 & 55.1/74.2 \\
 V2B & 70.5/81.3 & 48.3/73.5 & 50.1/58.0 & 40.8/49.7 & 58.4/75.2 \\
 CMT & 70.5/81.9 & 49.1/75.5 & 54.1/64.1 & 55.1/82.4 & 59.4/77.6 \\
 PTTR & 65.2/77.4 & 50.9/81.6 & 52.5/61.8 & 65.1/90.5 & 57.9/78.1 \\ 
 STNet & 72.1/\second{{84.0}} & 49.9/77.2 & 58.0/70.6 & 73.5/93.7 & 61.3/80.1 \\
 TAT & \second{{72.2}}/83.3 & 57.4/84.4 & 58.9/69.2 & 74.2/93.9 & 64.7/82.8\\ 
 M2-Track & 65.5/80.8 & 61.5/88.2 & 53.8/70.7 & 73.2/93.5 & 62.9/83.4 \\ 
 CXTrack & 69.1/81.6 & \second{{67.0}}/\second{91.5} & \second{60.0}/\second{71.8} & \second{74.2}/\second{\textbf{94.3}} & \second{67.5}/\second{85.3} \\
 \hline
 MBPTrack & \textbf{73.4}/\textbf{84.8} & \textbf{68.6}/\textbf{93.9} & \textbf{61.3}/\textbf{72.7} & \textbf{76.7}/\textbf{94.3} & \textbf{70.3}/\textbf{87.9} \\
 Improvement & \better{↑1.2}/\better{↑0.8} & \better{↑1.6}/\better{↑2.4} & \better{↑1.3}/\better{↑0.9} & \better{↑2.5}/0.0 & \better{↑2.8}/\better{↑2.6} \\
\hline
\end{tabular} }
\end{center}
\vspace{-1.0cm}
\end{table}

\begin{table*}
\begin{center}
\caption{\textbf{Comparison with state of the arts on Waymo Open Dataset.}}
\vspace{-0.3cm}
\label{waymo}
\resizebox{0.95\linewidth}{!}{
\begin{tabular}{c|cccc|cccc|c}
\hline 
\multirow{2}{*}{Method} & \multicolumn{4}{c|}{Vehicle(185731)} & \multicolumn{4}{c|}{Pedestrian(241752)} & \multirow{2}{*}{Mean(427483)}\\
& Easy & Medium & Hard & Mean & Easy & Medium & Hard & Mean & \\
\hline
 P2B& 57.1/65.4 & 52.0/60.7 & 47.9/58.5 & 52.6/61.7 & 18.1/30.8 & 17.8/30.0 & 17.7/29.3 & 17.9/30.1 & 33.0/43.8\\ 
 BAT& 61.0/68.3 & 53.3/60.9 & 48.9/57.8 & 54.7/62.7 & 19.3/32.6 & 17.8/29.8 & 17.2/28.3 & 18.2/30.3 & 34.1/44.4\\
 V2B& 64.5/71.5 & 55.1/63.2 & 52.0/62.0 & 57.6/65.9 & 27.9/43.9 & 22.5/36.2 & 20.1/33.1 & 23.7/37.9 & 38.4/50.1\\
 STNet& 65.9/72.7 & {{57.5/66.0}} & {{54.6/64.7}} & {{59.7/68.0}} & {29.2}/{45.3} & {24.7}/{38.2} & {22.2}/{35.8} & {25.5}/{39.9} & {40.4}/{52.1}\\ 
 TAT& 66.0/72.6 & 56.6/64.2 & 52.9/62.5 & 58.9/66.7 & 32.1/49.5 & 25.6/40.3 & 21.8/35.9 & 26.7/42.2 & 40.7/52.8\\
 CXTrack & 63.9/71.1 & 54.2/62.7 & 52.1/63.7 & 57.1/66.1 & 35.4/\second{55.3} & {29.7/47.9} & {26.3/44.4} & {30.7/49.4} & {42.2/56.7}\\
 M2Track & \second{{68.1}}/\second{75.3} & \second{\textbf{58.6}}/\second{66.6} & \second{55.4}/\second{64.9} & \second{{61.1}}/\second{69.3} & \second{35.5}/54.2 & \second{30.7}/\second{48.4} & \second{29.3}/\second{45.9} & \second{32.0}/\second{49.7} & \second{44.6}/\second{58.2}\\
\hline
MBPTrack & \textbf{68.5}/\textbf{77.1} & {58.4}/\textbf{68.1} & \textbf{57.6}/\textbf{69.7} & \textbf{61.9}/\textbf{71.9} & \textbf{37.5}/\textbf{57.0} & \textbf{33.0}/\textbf{51.9} & \textbf{30.0}/\textbf{48.8} & \textbf{33.7}/\textbf{52.7} & \textbf{46.0}/\textbf{61.0}\\

 Improvement & \better{↑0.4}/\better{↑1.8} & \worse{↓0.2}/\better{↑1.5} & \better{↑2.2}/\better{↑4.8} & \better{↑0.8}/\better{↑2.6} & \better{↑2.0}/\better{↑1.7} & \better{↑2.3}/\better{↑3.5} & \better{↑0.7}/\better{↑2.9} &  \better{↑1.7}/\better{↑3.0} & \better{↑1.4}/\better{↑2.8} \\
\hline
\end{tabular} }
\end{center}
\vspace{-0.6cm}
\end{table*}

\begin{table*}
\begin{center}
\caption{\textbf{Comparisons with the state-of-the-art methods on NuScenes dataset}. }
\vspace{-0.3cm}
\label{nuscene}
\resizebox{0.85\linewidth}{!}{
\begin{tabular}{c|c|c|c|c|c|c}
\hline 
Method & Car(64159) & Pedestrian(33227) & Truck(13587) & Trailer(3352) & Bus(2953) & Mean(117278) \\
\hline
 SC3D & 22.31/21.93 & 11.29/12.65 & 30.67/27.73 & 35.28/28.12 & 29.35/24.08 & 20.70/20.20 \\
 P2B & 38.81/43.18 & 28.39/52.24 & 42.95/41.59 & 48.96/40.05 & 32.95/27.41 & 36.48/45.08 \\ 
 BAT & 40.73/43.29 & 28.83/53.32 & 45.34/42.58 & 52.59/44.89 & 35.44/28.01 & 38.10/45.71 \\
 M2-Track & \second{55.85}/\second{65.09} & \second{32.10}/\second{60.92} & \second{57.36}/\second{59.54} & \second{57.61}/\second{58.26} & \second{51.39}/\second{51.44} & \second{49.23}/\second{62.73} \\ 
 \hline
 MBPTrack & \textbf{62.47}/\textbf{70.41} & \textbf{45.32}/\textbf{74.03} & \textbf{62.18}/\textbf{63.31} & \textbf{65.14}/\textbf{61.33} & \textbf{55.41}/\textbf{51.76} & \textbf{57.48}/\textbf{69.88}\\
 
 Improvement & \better{↑6.62}/\better{↑5.32} & \better{↑13.22}/\better{↑13.11} & \better{↑4.82}/\better{↑3.77} & \better{↑7.53}/\better{↑3.07} & \better{↑4.02}/\better{↑0.32} & \better{↑8.25}/\better{↑7.15}\\
\hline
\end{tabular} }
\end{center}
\vspace{-0.8cm}
\end{table*}

We present a comprehensive comparison of our method with the previous state-of-the-art approaches, namely SC3D~\cite{giancola2019leveraging}, P2B~\cite{qi2020p2b}, 3DSiamRPN~\cite{fang20203d}, LTTR~\cite{cui20213d}, MLVSNet~\cite{wang2021mlvsnet}, BAT~\cite{zheng2021box}, PTT~\cite{shan2021ptt}, V2B~\cite{hui20213d}, CMT~\cite{guo2022cmt}, PTTR~\cite{zhou2022pttr}, STNet~\cite{hui20223d}, TAT~\cite{lan2022tat}, M2-Track~\cite{zheng2022beyond} and CXTrack~\cite{xu2022cxtrack} on the KITTI dataset. The published results from corresponding papers are reported. As illustrated in Tab.~\ref{kitti}, MBPTrack surpasses other methods on all categories, with an obvious improvement in average Success and Precision. Notably, compared with point-based methods such as CXTrack or M2Track, methods using voxel-based localization heads like STNet and V2B achieve satisfying results on the Car category. We presume that the improvement stems from the simple shape and large size of cars, which fit well in voxels. However, STNet and V2B perform poorly on the Pedestrian category, which has small size and complex geometry. Voxelization results in inevitable information loss, causing the network to fail to distinguish the target from distractors. Leveraging box priors and a memory mechanism, our method achieves state-of-the-art performance on both categories. Compared with TAT, which samples high-quality target templates from historical frames, our method obtains consistent performance gains across all categories. It indicates that our method benefits a lot from spatial and temporal information that TAT discards during sampling. 

\begin{figure*}[htbp]
\centering
\includegraphics[width=1.0\linewidth]{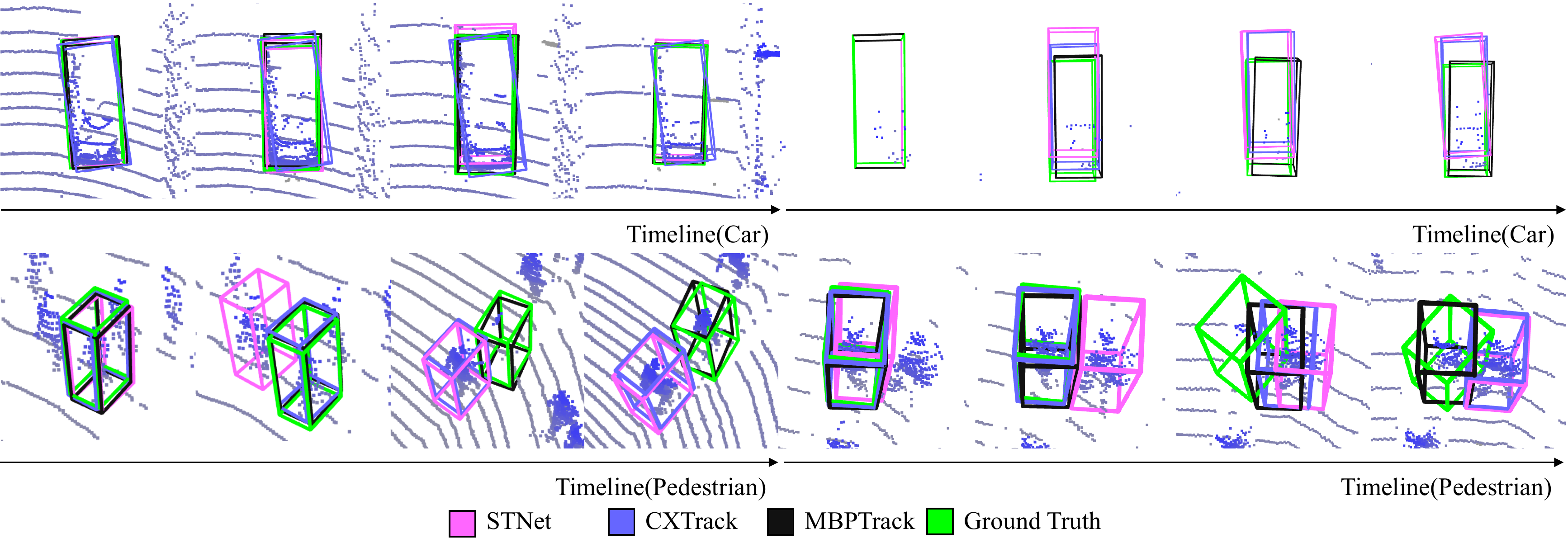}
\vspace{-0.7cm}
\caption{\textbf{Visualization of tracking results compared with state-of-the-art methods.}}
\label{fig:visual}
\vspace{-0.5cm}
\end{figure*}

We present a visual analysis of the tracking results on KITTI for further explanation. As shown in Fig.~\ref{fig:visual}, CXTrack~\cite{xu2022cxtrack}, which adopts a point-based head, fails to predict the orientation accurately on the Car category, while the predicted bounding boxes by our method hold tight to the ground truths. For pedestrians, all methods tend to drift towards intra-class distractors due to the large appearance variation caused by heavy occlusion. However, only MBPTrack can accurately track the target after the occlusion disappears, owing to the sufficient use of temporal information.

To demonstrate the generalization ability of our proposed MBPTrack, we evaluate the KITTI pretrained models on WOD~\cite{sun2020waymo}, following previous work~\cite{hui20223d}. The corresponding categories between KITTI and WOD datasets are Car→Vehicle and Pedestrian→Pedestrian.
The experimental results, as presented in Tab.~\ref{waymo}, indicate that MBPTrack yields competitive or better tracking results than other methods under different levels of sparsity. In conclusion, our proposed method not only precisely tracks targets of all sizes but also generalizes well to unseen scenarios.

NuScenes~\cite{caesar2020nuscenes} presents a greater challenge for 3D SOT task than KITTI due to its larger data volumes and lower frequency for annotated frames (2Hz for NuScenes v.s. 10Hz for KITTI and WOD). We conduct a comparison of our approach with previous methods on the NuScenes dataset following M2-Track~\cite{zheng2022beyond}. As shown in Tab.~\ref{nuscene}, our method achieves a consistent and large performance gain compared with the previous state-of-the-art method, M2-Track. Leveraging the rich temporal and spatial information contained in the historical frames, MBPTrack exhibits superior performance over methods that only consider two frames when large appearance variation occurs between them.

\begin{table}
\begin{center}
\caption{\textbf{Model complexity and inference time.}}
\vspace{-0.3cm}
\label{speed}
\resizebox{0.85\linewidth}{!}{
\setlength{\tabcolsep}{3pt}
\begin{tabular}{c|c|c|c}
\hline 
Component & FLOPs & \#Params & Inference Speed\\
\hline
backbone & 1.59G & 1.19M & 4.6ms\\
DeFPM & 0.22G & 2.67M & 7.9ms \\
BPLocNet & 1.07G & 3.52M & 3.6ms \\
pre/post-process & - & - & 3.9ms\\
\hline
MBPTrack & 2.88G & 7.38M & 20.0ms (50FPS)\\
\hline
\end{tabular} }
\end{center}
\vspace{-0.7cm}
\end{table}

Fig.~\ref{speed} shows the model complexity and average inference time of different components in the Car category on KITTI. Our experiments are conducted on a single NVIDIA RTX 3090. MBPTrack achieves 50 FPS, with 4.6ms for feature extraction, 7.9ms for feature propagation, 3.6ms for localization and 3.9ms for pre/post-processing. Using a more sophisticated attention mechanism (e.g.~\cite{lee2019set,katharopoulos2020transformers}) in DeFPM may further increase the running speed.

\subsection{Ablation Studies}

\label{sec:ablation}

\begin{table}
\begin{center}
\caption{\textbf{Ablation study of the memory size.}}
\vspace{-0.3cm}
\label{memsize}
\resizebox{\linewidth}{!}{
\setlength{\tabcolsep}{3pt}
\begin{tabular}{c|c|c|c|c|c}
\hline 
\#Frames & Car & Pedestrian & Van & Cyclist & Mean\\
\hline
1 & 72.5/83.7 & 63.2/88.9 & \textbf{62.8}/\textbf{74.3} & 74.6/93.7 & 67.7/85.3 \\
2 & 72.5/82.8 & 66.8/91.9 & 62.0/73.4 & 76.6/94.4 & 69.2/86.2 \\
3 (Ours) & 73.4/84.8 & \textbf{68.6}/\textbf{93.9} & 61.3/72.7 & 76.7/94.3 & \textbf{70.3}/\textbf{87.9} \\
4 & \textbf{74.9}/\textbf{86.7} & 66.4/92.1 & 60.7/72.0 & 76.7/94.4 & 70.0/\textbf{87.9} \\
5 & 74.1/85.5 & 66.7/92.2 & 61.2/72.4 & 76.7/94.4 & 69.8/87.4 \\
6 & 72.2/83.5 & 65.6/91.0 & 60.1/71.5 & \textbf{77.4}/\textbf{94.6} & 68.4/85.9 \\
\hline
\end{tabular} }
\end{center}
\vspace{-1.0cm}
\end{table}

\noindent\textbf{Memory Size.} Memory size is defined as the number of historical frames with their corresponding targetness masks saved in the memory. To explore the impact of memory size, we conduct experiments on KITTI and report the results in Tab.~\ref{memsize}. Notably, we train and test our model using only one previous frame when the memory size is set as $1$. In this case, MBPTrack is degraded to a Siamese-based network, same as previous work~\cite{hui20223d,xu2022cxtrack,zheng2022beyond}. Otherwise, we adopt the default training settings. Compared with Siamese-based version, our method benefits a lot from exploiting temporal information, leading to a significant improvement on the average metrics. This demonstrates the importance of historical information. We also observe that performance begins to decline when using more than 4 frames for tracking. Larger memory size can provide more reference information to handle sudden appearance variation caused by occlusion, but may fail to tackle lasting appearance variation, such as the gradual sparsification of point clouds as the tracking target moves further away. Besides, the memory size at which the model's performance reaches its peak varies across different categories. We believe that the peak point is determined by the quality of point clouds. For example, on the Car category, the tracked target may suffer from heavy occlusion and data missing, and thus it requires a large memory size to capture much shape information.

\begin{table}
\begin{center}
\caption{\textbf{Ablation studies of different model components.} ``De'', ``Q'' and ``PS'' denote the decoupling design in DeFPM, coarse-to-fine score prediction and positive sampling strategy for non-rigid objects, respectively.}
\label{ablation_comp}
\vspace{-0.3cm}
\resizebox{\linewidth}{!}{
\begin{tabular}{c|c|c|c|c|c|c|c}
\hline 
De & Q & PS & Car & Pedestrian & Van & Cyclist & Mean\\
\hline
 
 & \checkmark & \checkmark & 70.0/81.3 & 64.1/88.5 & 58.7/70.4 & 72.5/93.4 & 66.5/83.7\\
 \checkmark & & \checkmark & 71.8/82.9 & 64.2/89.5 & 59.0/69.5 & 74.9/93.9 & 67.4/84.8\\
 \checkmark & \checkmark & & 73.4/84.8 & 65.6/91.6 & 61.3/72.7 & 75.1/94.0 & 69.0/86.9\\
\checkmark & \checkmark & \checkmark & \textbf{73.4}/\textbf{84.8} & \textbf{68.6}/\textbf{93.9} & \textbf{61.3}/\textbf{72.7} & \textbf{76.7}/\textbf{94.3} & \textbf{70.3}/\textbf{87.9}\\
\hline
\end{tabular} }
\end{center}
\vspace{-0.7cm}
\end{table}

\noindent\textbf{Model components.} Tab.~\ref{ablation_comp} presents ablation studies of MBPTrack on KITTI to gain a better understanding of its model designs. We investigate the impact of decoupling design in DeFPM, coarse-to-fine targetness score prediction in BPLocNet and positve sampling training strategy for non-rigid objects like pedestrians and cyclists via separate ablation experiments. Although the effectiveness of different components varies across categories, removing any of them leads to an obvious decline in terms of average metrics.

\begin{table}
\begin{center}
\caption{\textbf{Ablation studies of different localization heads.}}
\label{ablation_loc}
\vspace{-0.3cm}
\resizebox{\linewidth}{!}{
\begin{tabular}{c|c|c|c|c|c}
\hline 
Loc & Car & Pedestrian & Van & Cyclist & Mean\\
\hline
 RPN & 67.2/81.1 & 53.5/85.5 & 52.0/62.4 & 61.3/90.2  & 59.8/81.5\\
PRM & 69.0/81.4 & 59.0/88.4 & 54.0/64.5 & 71.6/92.6 & 63.4/83.2 \\
V2B & 72.6/84.2 & 61.1/87.9 & 55.6/64.7 & 71.2/93.8 & 66.1/84.3\\
X-RPN & 70.4/81.9 & 64.9/91.3 & 55.1/64.6 & 72.1/93.2 & 64.7/84.7\\
BPLoc & \textbf{73.4}/\textbf{84.8} & \textbf{68.6}/\textbf{93.9} & \textbf{61.3}/\textbf{72.7} & \textbf{76.7}/\textbf{94.3} & \textbf{70.3}/\textbf{87.9}\\
\hline
\end{tabular} }
\end{center}
\vspace{-1.0cm}
\end{table}

\begin{table}
\begin{center}
\caption{\textbf{Integration with Siamese-based network.}}
\vspace{-0.3cm}
\label{integration}
\resizebox{\linewidth}{!}{
\setlength{\tabcolsep}{3pt}
\begin{tabular}{c|c|c|c|c|c}
\hline 
Method & Car & Pedestrian & Van & Cyclist & Mean\\
\hline
CXTrack & 69.1/81.6 & 67.0/91.5 & 60.0/71.8 & \textbf{74.2}/\textbf{94.3} & 67.5/85.3 \\
CXTrack${}^\dagger$ & \textbf{72.8}/\textbf{84.5} & \textbf{67.7}/\textbf{92.1} & \textbf{61.3}/\textbf{72.7} & 74.1/94.1 & \textbf{69.6}/\textbf{87.0} \\
Improvement & \better{↑3.7}/\better{↑2.9} & \better{↑0.7}/\better{↑0.6} & \better{↑1.3}/\better{↑0.9} & \worse{↓0.1}/\worse{↓0.2} & \better{↑2.1}/\better{↑1.6}\\
\hline
\end{tabular} }
\footnotesize{$\dagger$: integrated with BPLocNet} 
\end{center}
\vspace{-1.0cm}
\end{table}

\noindent\textbf{Localization head.} We compare our proposed BPLocNet and other commonly-adopted localization heads on KITTI, including point-based (RPN~\cite{qi2019deep}, PRM~\cite{zhou2022pttr}, X-RPN~\cite{xu2022cxtrack}) and voxel-based (V2B~\cite{hui20213d}) methods. The results are shown in Tab.~\ref{ablation_loc}. BPLocNet consistently outperforms the alternative designs on all categories. We further integrate BPLocNet with a previous Siamese-based method CXTrack~\cite{xu2022cxtrack} to explore its generalization ability. Tab.~\ref{integration} shows obvious performance gain by using BPLocNet, especially on the Car category (72.8/84.5 v.s. 69.1/81.6). For cars that have simple shapes and suffer from self-occlusions, box-prior sampling provides a strong shape prior to the localization task, thereby leading to better performance than the point-based X-RPN adopted in CXTrack~\cite{xu2022cxtrack}.

\section{Conclusion}

We propose a memory-based tracker, named MBPTrack, to address 
the appearance variation and size difference problems in 3D single object tracking. MBPTrack employs a decoupling feature propagation module to exploit rich information lying in historical frames, which is overlooked by previous Siamese-based methods. We also design a novel localization network, named BPLocNet, that leverages box priors to more accurately localize the tracked targets of different sizes. Extensive experiments on three large-scale datasets show our method surpasses previous state-of-the-art on tracked targets of varying sizes while maintaining high efficiency. The major limitation of our work is the inaccurate orientation prediction caused by inaccurate past predictions (Fig.~\ref{fig:visual}, bottom right). Besides, our method achieves limited performance when the point cloud is extremely sparse. In the future, we would like to explicitly model the target motion to address these issues.

{\small
\bibliographystyle{ieee_fullname}
\bibliography{egbib}
}

\end{document}